\documentclass[a4,12pt]{article}
\usepackage{latexsym}
\usepackage{epsfig}
\usepackage{amsmath,amstext,amsfonts}

\makeatletter
  
  \@addtoreset{equation}{section}
\makeatother 
\title{Multiplicative  Algorithm for  Orthgonal Groups \\and 
Independent Component Analysis }
\author{Toshinao {\sc
    Akuzawa}\thanks{akuzawa@brain.riken.go.jp}\vspace{0.3cm}\\
\vspace{0.5cm}\\
Brain Science Institute \\
{\it RIKEN}\\
{\small 2-1 Hirosawa, Wako, Saitama 351-0198, Japan}}
\date{{\it \today}}

\begin{document}
\maketitle
\abstract{
The multiplicative Newton-like method developed by the author\:{\it et\:al.}
is extended to
the situation  where the dynamics  is restricted  to
 the orthogonal group. 
A general framework is
constructed  without specifying the cost function. 
Though the restriction to the orthogonal groups   makes the problem 
 somewhat complicated, 
an explicit expression for the amount of  individual jumps is obtained. 
This algorithm is exactly second-order-convergent. 
The global instability inherent in the Newton method is remedied by
 a Levenberg-Marquardt-type variation.  
The method thus constructed  can readily be applied to the independent
component analysis. 
Its remarkable performance is illustrated by a
numerical simulation.

% In the case of the independent component analysis  the restriction
% corresponds to the prewhitening of the  data. 
 }
\section{Overview}
\label{intro}
Many optimization problems take the form, 
``Find an optimal matrix under the constraints (1).. (2).. {\it etc}."
Some of these can be considered as optimizations on Lie groups. 
For groups, the fundamental manipulation
is a multiplication whereas an addition is unnatural. 
%(Imagine the compound interest rate on your bank account.)
In consideration of this fact, 
we have constructed a multiplicative Newton-like algorithm 
for maximizing the kurtosis (a good barometer for the independence) in 
\cite{akuzawa8}.  There the dynamics takes place on the coset 
$GL(1,{\Bbb R})^{N}\backslash GL(N,{\Bbb R})$. 
We can apply the techniques
developed in \cite{akuzawa8} to many other optimization problems. 
The coset structure $GL(1,{\Bbb R})^{N}\backslash GL(N,{\Bbb R})$ is,
however,
characteristic of the  independent component
analysis(ICA). It is understood 
by the fact that the independence is nothing to do with the scaling. 
The redundancy
resulting from the invariance of the model under the componentwise scaling 
must be eliminated for a rigorous discussion and this redundancy
corresponds 
to $GL(1,{\Bbb R})^{N}$. 

Another way to eliminate this redundancy is the
prewhitening. 
The prewhitening is a linear transformation of the observed data 
which  maps
the covariance matrix to  the unit matrix. 
If we deal with  prewhitened data, we can legitimately narrow
the sweeping range  to the orthogonal group. 
The aim of this letter is the construction of a multiplicative
algorithm
for the orthogonal groups.

The framework is  as follows. 
%Suppose that 
 $N$-dimensional prewhitened random variables 
 $\{X_i|1\le i \le N\}$ are available
and it is anticipated that their origins  are 
 some unknown mutually independent components $\{Y_i^*|1\le i \le N\}$.
The goal of the ICA is the map
 $\{X_i\} \mapsto \{Y_i^*\}$. 
We restrict ourselves to 
 the linear independent component analysis. 
There  
we want to find a linear transformation $C^*:{X}=(X_1,\cdots,X_N)'\mapsto
{ Y^*}=(Y_1^*,\cdots,Y_N^*)'=C^*{ X}$ which 
 minimizes some cost function that measures the independence. 
Since we are assuming that the data is already prewhitened, the
covariance matrix of $X$ is the $N\times N$ unit matrix.
If we do not take into account  errors in the prewhitening, 
the optimal  point $C^*$ must belong to $O(N)$.

Giving up the analytical solution, 
we consider a sequence, 
\begin{eqnarray}
  \label{eq:intro1}
  C(0),~ C{(1)},~ C{(2)},~ C{(3)},~\cdots\cdots~, 
\end{eqnarray}
 which converges to the optimal solution $C^*$.  
The sequence  $\{C(t)\}$ 
% which specifies the
%linear transformation 
is generated by the left-multiplication of another sequence of
orthogonal  
matrices $\{D(t)\}$. 
Each $D(t)$  is specified by the coordinate 
$\Delta(t)$ which satisfies $D(t)={\rm e}^{\Delta(t)}$. 
We assume that $\Delta(t)$ is 
an $N\times N$  skew-symmetric 
matrix,  
which  implies that  $D(t)$ belongs to
the identity component of $O(N)$. 
In practice the procedure  is as follows.
As an initial condition we set $C(0)$. 
For  $t>0~(t\in{\Bbb N}^{+})$, 
we introduce %an  $N\times N$ matrix 
$\Delta(t)$ and 
denote $C({t})$ as  $C({t+1})={\rm e}^{\Delta({t})}C(t)$.
Under these settings, we determine $\Delta(t)$ by using the Newton method 
%for the second order
%expansion of the cost function 
with respect to 
the matrix elements of 
$\Delta(t)$. That is,
 we evaluate the cost function at $C({t+1})$ 
by  expanding it  around $C({t})$ 
in terms of  the elements of
$\Delta({t})$ up to the second order. 
 Then    $\Delta(t)$ is choosen as the (unique) critical point of 
this second order
expansion.  
We iteratively follow these procedures until we obtain a satisfactory
solution. 

This letter is organized as follows. 
In Section \ref{sec:mult} 
we will give a complete description of 
a new  multiplicative updating method for the orthogonal groups. 
This section  is the main part of this letter. Since our formulation 
does not depend on the details of the  cost function
the method can be useful for many problems other than the ICA. 
The performance of
our method including the second-order-convergence is discussed in
Section \ref{sec:per1}.   
Section \ref{sec:appl} is a survey of possible applications of our
method. 
The algorithm constructed in Section \ref{sec:mult} 
is considered as  a pure-Newton method on the orthogonal groups.
To achive  the global convergence, we must modify the method. This is 
accomplished  in 
Section \ref{sec:practice}. Section 
\ref{sec:practice} also includes a numerical examination of 
the performance of our
method. Section \ref{sec:summ} is a summary. 

\section{Multiplicative updating on $O(N)$}
\label{sec:mult}
We assume that the  cost function $F$ takes the form, 
\begin{eqnarray}
  \label{eq:a1}
  F(Y)=\sum_{i=1}^NE(f_i(Y_i))~,
\end{eqnarray}
where each $f_i:{\Bbb R}\rightarrow{\Bbb R}$ is an unspecified function. 
Through this letter we denote by $E(\cdot)$ the expectation.  % of $A$. 
We will determine 
the concrete procedures 
%amount of  each step 
 after  the Newton manner.  
First, we  introduce
 maps, 
 $R$ and $\{U_{i}(1\le i\le N)\}$'s,  from 
$N$-dimensional 
dataset to  $N \times N$ matrices
by 
\begin{eqnarray}
  \label{eq:a2}
  [R(Y)]_{ki}=E\left(\frac{\partial f_i(Y_i)}{\partial Y_i}Y_k\right)
\end{eqnarray}
and
\begin{eqnarray}
  \label{eq:a3}
[ U_{i}(Y)]_{kl}=U_{ikl}(Y)= E\left(\frac{\partial^2 f_i(Y_i)}{\partial
  Y_i^2}Y_k Y_l\right)~.
\end{eqnarray}
The goal is  the construction of  a sequence
$\{Y(t)\}$  of the estimates of the independent components, which
converges to the optimal point $Y^*$.  
%We suppose that
Within the framework of the linear analysis, we consider that 
 this sequence is derived from another sequence 
 $\{C(t)\}$ of the linear transformation by the relation
$Y(t)=C(t)X$,  
where $X$ are the original data. Thus if we  restate the problem,
 the task is to 
determine 
a  sequence  $\{C(t)\}$. 
We assume that
for each $t\in {\Bbb N}^{+}$
 the estimates of the independent components at  time $t$ and 
and the estimates 
at time $t+1$ are related by 
\begin{eqnarray}
  \label{eq:a4}
  Y{(t+1)}=D{(t)}Y{(t)}~
\end{eqnarray}
or equivalently
\begin{eqnarray}
  \label{eq:a4bb}
  C{(t+1)}=D{(t)}C{(t)}~,
\end{eqnarray}
where $D{(t)}$  is  some orthogonal matrix to be fixed. 
Our method is characterized by this left-multiplicative updating rule. 
As mentioned in the previous section,
we  assume that 
each $D(t)$   always belongs to the identity component of the 
orthogonal group $O(N)$. 
This assumption is reasonable, for example, if the  original data $X$
are already prewhitened in the case of the ICA. 
% we suppose that the original data $X$ are already prewhitened. 
%In this case  we can legitimately
Anyway, under this restriction 
 $D{(t)}$ is specified by an $N\times N$ anti-symmetric matrix $\Delta{(t)}$,
which satisfies
\begin{eqnarray}
  \label{eq:a5}
  \exp(\Delta{(t)})=D{(t)}~.
\end{eqnarray}
For brevity's sake we will omit the argument $(t)$ and denote $Y(t+1)$ by $Z$. 
$F(Z)$ is expanded in terms of $\{\Delta_{ij}\}$ as 
\begin{eqnarray}
  \label{eq:a6}
  F(Z)=F(Y)+{\rm tr}(\Delta R(Y))+{\rm
  tr}\left(\frac{\Delta^2}{2}R(Y)\right)
+\frac{1}{2}\sum_{i,k,l}\Delta_{ik}\Delta_{il}U_{ikl}(Y)
+O(\Delta^3)~. 
\end{eqnarray}
%By partially differentiating (\ref{eq:a6}),  
Through the letter
 we denote by $O(\Delta^k)$ polynomials of matrix elements of $\Delta$ 
which does not contain terms with degrees less than $k$. Do not
 confuse this with the symbol for the orthogonal groups such as  $O(N)$. 
As in the usual Newton method, 
we truncate the expansion (\ref{eq:a6}) at the second order with
respect to $\{\Delta_{ij}\}$.
 Then $\Delta$ in this step is determined as the coordinate of  the
critical point of this truncated expansion. 
The partial derivative of (\ref{eq:a6}) is more convenient for the purpose. 
It reads
\begin{eqnarray}
  \label{eq:a7}
  \frac{\partial F(Z)}{\partial \Delta_{kl}}
=R_{lk}+\frac{1}{2}\left[\Delta R+R\Delta
\right]_{lk}+\sum_p \Delta_{kp}U_{klp}+O(\Delta^2)~, 
\end{eqnarray}
where we have omitted the argument $Y$ for $R$ and $U$. 
Now let us  introduce a map $\rm cs$ (the column string) as in the previous
article
\cite{akuzawa8}:
\begin{eqnarray}
  \label{eq:a14}
  {\rm Mat}(N,{\Bbb F}) &\rightarrow& {\Bbb F}^{N^2}\\
A=\left(
  \begin{array}{cccc}
 A_{11}& A_{12}&\cdots &A_{1N}\\
A_{21} &\multicolumn{3}{c}{\dotfill}\\
\multicolumn{4}{c}{\dotfill}\\
A_{N1} &\multicolumn{2}{c}{\dotfill}&A_{NN}
  \end{array}
\right) &\mapsto& 
{\rm cs}(A)=
(A_{11}~ A_{21}~ \cdots~ A_{N1}~~ A_{12}~ A_{22}~\cdots ~A_{NN})'~,\nonumber
\end{eqnarray}
where  ${\rm Mat}(N,{\Bbb F})$ is  
 $N\times N$ matrices on some unspecified field $\Bbb F$.
We  denote by the upper subscript $\prime$ the transposition and
by $\dagger$ the complex conjugate. 
For the orthogonal groups it is rather simple to move to
the framework of the column string as compared to the case of
$GL(1,{\Bbb R})^N\backslash GL(N,{\Bbb R})$: 
By neglecting  $O(\Delta^2)$ terms, 
the right-hand-side of (\ref{eq:a7}) is straightforwardly  rewritten as 
\begin{eqnarray}
  \label{eq:a8}
R_{lk}&+&\frac{1}{2}\left[\Delta R+R\Delta
\right]_{lk}+\sum_p \Delta_{kp}U_{klp}\nonumber\\
&&=\left[{\rm cs}(R)+\frac{1}{2}\left(
R'\otimes I_N+I_N\otimes R
\right) {\rm cs}(\Delta)+
\big(\bigoplus_k U_k\big) T{\rm cs}(\Delta)\right]_{l+(k-1)N}~,
\end{eqnarray}
where
 the symbol ``$\bigoplus$'' stands for the direct sum,
\begin{eqnarray}
  \label{eq:tiu1}
\bigoplus_{k=1}^N U_{k}=
\left(
  \begin{array}{lllll}
U_1 & 0 & \multicolumn{2}{c}{\cdots\cdots} & 0\\ 
0& U_2 & 0 & \multicolumn{2}{c}{\cdots\cdots}\\
 \multicolumn{5}{c}{\dotfill}\\
 \multicolumn{5}{c}{\dotfill}\\
 0& \multicolumn{2}{c}{\cdots\cdots}& U_{N-1}& 0   \\
0& 0& \multicolumn{2}{c}{\cdots\cdots}& U_{N}   \\
   \end{array}
\right)~,
\end{eqnarray}
 $T$ is  an $N^2\times N^2$ matrix 
defined by 
\begin{eqnarray}
  \label{eq:a15}
  {\rm cs}(A')=T{\rm cs}(A) ~\mbox{\rm for~} A\in  {\rm Mat}(N,{\Bbb F})~,
\end{eqnarray}
and $I_N$ is the $N\times N$ unit matrix. 
We denote  the tensor product by  $\otimes$  as usual. 
 The ``transposition'' $T$ is also considered as 
an intertwiner between  two equivalent representations: 
\begin{eqnarray}
%\nonumber  
T(A\otimes B)T=B\otimes A~.
\end{eqnarray}
The orthogonal group $O(N)$ has less degrees of freedom than the
general linear group. 
The canonical basis of the Lie algebra, ${\frak o}(N)$, of $O(N)$ is
$N(N-1)/2$
anti-symmetric
matrices. We will introduce some operators which enable us to move to
the coordinates based on  
the canonical basis on ${\frak o}(N)$. 
In the first place, we introduce an $N^2\times N^2$ matrix  $H$ by
\begin{eqnarray}
  \label{eq:a9}
H=\sum_{i>j}H^{(i,j)}~,
\end{eqnarray}
where $H^{(i,j)}$ is a $\pi/4$  rotation between 
 the $j+N(i-1)$-th component and the $i+N(j-1)$-th component:
\begin{eqnarray}
  \label{eq:a10}
H^{(i,j)}_{kl}=  \left\{
  \begin{array}{ccl}
\frac{1}{\sqrt{2}}~~~~&\mbox{\rm for}&k=j+N(i-1),~~l=j+M(i-1)\\
-\frac{1}{\sqrt{2}}~~~~&\mbox{\rm for}&k=j+N(i-1),~~l=i+M(j-1)\\
\frac{1}{\sqrt{2}}~~~~&\mbox{\rm for}&k=i+N(j-1),~~l=j+M(i-1)\\
\frac{1}{\sqrt{2}}~~~~&\mbox{\rm for}&k=i+N(j-1),~~l=i+M(j-1)\\
0~~~~&&\mbox{\rm otherwise. }
  \end{array}
\right. 
\end{eqnarray}
The projection  operator $P_D$,
\begin{eqnarray}
  \label{eq:a18}
P_D&=&{\rm diag}(p_1,\cdots,p_{N^2})~,\nonumber\\
&&\left\{
\begin{array}{ll}
 p_k=1 ~~~\mbox{\rm for}~~ k=N(i-1)+i,1\le i\le N~\\
 p_k=0~~~~ \mbox{\rm otherwise}~,
\end{array}
\right.
\end{eqnarray}
 is used to extract the diagonal
elements of a matrix from its image by $\rm cs$. 
Then the coordinate transformation 
is realized by a multiplication of
\begin{eqnarray}
  \label{eq:a12.1}
  H+P_D~
\end{eqnarray}
to   column string vectors. 
We need to introduce two more 
projection operators $P_S$ and $P_A$  defined by
\begin{eqnarray}
  \label{eq:a11}
  P_S&=&{\rm diag}(p_1,p_2,\cdots,p_{N^2})\\
P_A&=&{\rm diag}(1-p_1,1-p_2,\cdots,1-p_{N^2})~,%\nonumber
\end{eqnarray}
where
\begin{eqnarray}
  \label{eq:a12}
  p_k=\left\{
    \begin{array}{ccl}
1&\mbox{\rm if}&{}^{\exists}(i,j);~~ j\le i~~ \mbox{\rm and}~~k=i+N(j-1)\\
0&&\mbox{\rm otherwise}. 
    \end{array}
\right.
\end{eqnarray}
By the left-action of  $P_S$ and $P_A$ to
 column string vectors rotated by $H+P_D$
we can extract,  
%The projection operators  $P_S$ and $P_A$  are used to extract,
respectively, 
 the symmetric components
 and the anti-symmetric components of the matrices.
Then the conditions 
for the critical point of the second-order-expansion,
which must be  satisfied by $\Delta$,  are 
translated into the following two conditions.
First, symmetric components of $\Delta$ must vanish.
 This condition is expressed as 
\begin{eqnarray}
  \label{eq:11.91}
\left[(H+P_D){\rm cs}(\Delta)\right]_{j+(i-1)N}=0
\qquad\mbox{\rm for}\quad i\le j
\quad\bigg(\Longleftrightarrow
P_S(H+P_D){\rm cs}(\Delta)  =0\bigg)~.
\end{eqnarray}
Secondly, for the anti-symmetric components 
the condition for the critical point is transformed to 
\begin{eqnarray}
  \label{eq:a11.9}
  \left[(H+P_D){\rm cs}(R)+(H+P_D)W
 {\rm cs}(\Delta)
\right]_{j+(i-1)N}~=0 \qquad\mbox{\rm for}\quad i>j~,
\end{eqnarray}
where we have  set 
\begin{eqnarray}
  \label{eq:a14b}
  W=\frac{1}{2}\left(
R'\otimes I_N+I_N\otimes R
\right) +
\big(\bigoplus_k U_k\big) T~.
\end{eqnarray}
%Now 
%one can see that 
% (\ref{eq:a8}) 
The conditions (\ref{eq:11.91}) and (\ref{eq:a11.9}) are 
combined into an equation, 
\begin{eqnarray}
  \label{eq:a13-1}
P_A(H+P_D){\rm cs}(R)+
 \bigg[P_A (H+P_D) W (H+P_D)' P_A +P_S
\bigg](H+P_D) {\rm cs}(\Delta)
=0~.
\end{eqnarray}
Note that 
\begin{eqnarray}
  \label{eq:a12.2}
  P_A(H+P_D)=P_AH~.
\end{eqnarray}
The optimal $\Delta$ is immediately obtained from (\ref{eq:a13-1}): 
\begin{eqnarray}
  \label{eq:a13}
  {\rm cs}(\Delta)&=&
-(H+P_D)'\bigg[P_A (H+P_D) W (H+P_D)' P_A +P_S
\bigg]^{-1}P_A(H+P_D){\rm cs}(R)~\nonumber\\
&=&
-H'\left(P_A H W H' P_A +P_S
\right)^{-1}P_AH{\rm cs}(R)~.
\end{eqnarray}
Thus we have obtained the explicit updating rule. 
By iterating the procedure in this section  from a  starting point 
sufficiently close
to the 
optimal one, 
 the sequences  $\{C(t)\}$ and $\{Y(t)\}$ converge to 
 the optimal solutions. 

\section{Performance (theoretical aspects)}\label{sec:per1}
%Our method has  very desirable convergence properties. 
The second-order-convergence is one of the main  advantages of this
method.
Indeed, this algorithm is rigorously  second-order-convergent. The
proof   can be  given 
almost in the same way as in \cite{akuzawa8}. So we omit the proof in
this letter.  

Sometimes we have to
deal with  large matrices  to apply  the technique here constructed. 
Let us examine the situation. 
The $N^2\times N^2$ matrix $P_A HW H' P_A +P_S$ is 
a direct sum of an $N(N-1)/2\times N(N-1)/2$ matrix
and an  $N(N+1)/2\times N(N+1)/2$ unit matrix. 
Within the   $N(N-1)/2\times N(N-1)/2$
block 
the number of non-zero off-diagonal elements   is 
no more than  ${N(N-1)(N-2)}$.  
So this is a very sparse matrix when $N$ becomes large. 
Of course if $N$ becomes extremely large, our method requires quite large
memories. But due to the sparseness, it remains to be  a 
practical tool for problems with considerably large $N$. 
\begin{figure}[htbp]
  \begin{center}
\epsfig{file=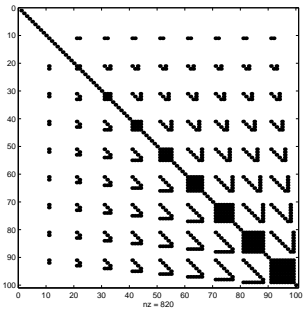, scale=0.3}
\caption{\small  $N=10$. The black dots denote non-zero elements of $P_A H W H' P_A +P_S$.  }
      \end{center}
\end{figure}

As  is often the case with the Newton method,  % \cite{akuzawa8}
the global convergence is not assured by this algorithm.
%So first few steps must be treated separately. 
Fortunately it is possible to  cure this fault. 
We will show the prescription to the global instability in
Section \ref{sec:practice}. 

%it is not assured that  this method 
%converges  globally.

\section{Applications to ICA}\label{sec:appl}
So far we have not specified the cost function beyond the assumption
that 
the cost function is a sum of the form (\ref{eq:a1}). 
Many of the cost functions  for the independent component analysis
  belong to this class. 
\subsection{Kullback-Leibler information}
The Kullback-Leibler information,
 \begin{eqnarray}
  \label{eq:ka9}
\int \prod_{i=1}^Ndy_i P(y)\bigg\{\ln  P(y)- \sum_{i=1}^N \ln
P_i(y_i)\bigg\}
~,
\end{eqnarray}
is a good measure for the independence. 
 Here $P$ is the joint probability density function of $\{Y_i\}$ and 
 $P_i$ is the probability density function of the $i$-th component. 
We have already restricted ourselves to the case where the jacobian of
 the transformation equals one. Then 
the minimization of the Kullback-Leibler information  is equivalent to 
  the minimization of 
  \begin{eqnarray}
    \label{eq:bb1.1}
 -\int \prod_i dY_i P(Y)\sum_{i=1}^N\ln  P_i(Y_i)
=\sum_{i=1}^N E(-\ln P_i(Y_i)) ~ .     
  \end{eqnarray}
Thus we can  legitimately  
transform 
the Kullback-Leibler information  
to a cost
function of the
form  (\ref{eq:a1}), where we
  should set $\{f_i\}$'s as 
\begin{eqnarray}
  \label{eq:bb1}
  f_i(\cdot)= -\ln  P_i(\cdot)~.
\end{eqnarray}
We must evaluate $\{P_i\}$'s,  their derivatives, and so on  to determine
the optimal
solution. A robust estimation 
of these quantities  is possibly  not an easy  task\cite{silverman1,cox1}. 

\subsection{Cumulant of fourth order}\label{subsec:cum}
The kurtosis of a random variable $A$ is defined by 
  \begin{eqnarray}
    {\kappa(A)}
=\frac{E(A^4)}{(E(A^2))^2}-3~.
  \end{eqnarray}
The kurtosis is related to the cumulant of the fourth order, 
\begin{eqnarray}
%  \nonumber
Cum^{(4)}(A)=E(A^4)-3(E(A^2))^2~,
\end{eqnarray}
by
  \begin{eqnarray}%\nonumber
    {\kappa(A)}=\frac{    Cum^{(4)}(A)}{(E(A^2))^2}~.
  \end{eqnarray}
For prewhitened data the kurtosis equals the cumulant of the fourth
order. 
As is well-known\cite{hyvarinen1,akuzawa8}, 
we can grab  independent components in many cases
by seeking  the maximum of the absolute values of the kurtoses. Our method
is applicable 
by setting
\begin{eqnarray}
  \label{eq:kur1}
 f_i=-\kappa^2  
\end{eqnarray}
for all $i$. 
If it is  known a priori that all the sources $\{Y_i^*\}$ have positive
kurtoses, we may use the kurtosis itself and  set 
\begin{eqnarray}
  \label{eq:kur2}
 f_i=-\kappa~.  
\end{eqnarray}
For these cost functions, $R$, $\{U_i\}$, and other
quantities needed for determining each step are calculated easily
from the observed data.
Thus applying our method for this cost function is highly practical and 
reasonable choice. 
\section{Levenberg-Marquardt-type variation and performance in practice}
\label{sec:practice}
The pure-Newton updating rule (\ref{eq:a13}) has a 
poor global convergence property.
This drawback is remedied  by
 the Levenberg-Marquardt-type variation\cite{numerical1}. 
First, We modify  (\ref{eq:a13}) 
as 
 \begin{eqnarray}
  \label{eq:lev1}
  {\rm cs}(\Delta)&=&
-H'\left(P_A H W H' P_A +P_S+\lambda I_{N^2}
\right)^{-1}P_AH{\rm cs}(R)~.
\end{eqnarray}
The initial value $\lambda_0$ for $\lambda$ is fixed at some positive value.
We also fix a real number  $\alpha(>1)$. 
(In the following example we set $\lambda_0=50$ and $\alpha=10$.) 
Then the  procedure at time $t$ is as follows:
\renewcommand{\labelenumi}{\roman{enumi})}
\begin{enumerate}
\item 
Calculate $\Delta$ by  (\ref{eq:lev1}). 
\item 
If $F({\rm e}^{\Delta}Y(t))$ is larger than $F(Y(t))$,
multiply $\lambda$ 
by $\alpha$ and go back to i). 
\item 
Otherwise, 
multiply $\lambda$ by $1/\alpha$ and proceed to the next time step $t+1$. 
\end{enumerate}
Other parts of the algorithm is completely the same  as in the
pure-Newton version in Section \ref{sec:mult}.

\begin{figure}[htbp]
  \begin{center}
\epsfig{file=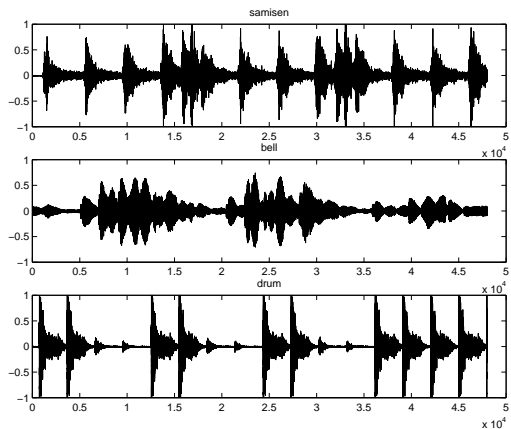,scale=0.4}    
    \caption{Sample  data generated by a synthesizer (by courtesy of 
N.Murata).}
    \label{fig:2}
  \end{center}
\end{figure}

Let us  examine the real performance of 
our method under this setting.
For the cost function we choose the kurtosis as in Subsection
\ref{subsec:cum}. 
 The source signals are three synthesizer-generated 
wav files(Fig.\ref{fig:2}). 
Pseudo-observed data  are generated by mixing the source by 
a random  matrix,
 \begin{eqnarray}
   \label{eq:tr1}
   A=I_3+S,
 \end{eqnarray}
where each element of $S$ is distributed uniformly on $(-1/2,1/2)$. 
 The residual crosstalk of the  signals 
demixed by our method 
is 
$1.29\%$ on average.  It takes about $122$ seconds (CPU time) for one
 hundred iteration of the same problem on
 our workstation. 
For reference, we have also solved the same demixing problem
by the FastICA\cite{fastica1}. 
In this case the residual crosstalk 
is 
$1.36\%$ on average and  it takes about $156$ seconds for 
one hundred 
 iteration on
 the same workstation. 
Since the author's  knowledge about the FastICA package is limited, 
one should not take this result seriously. 
It can, however, be said 
 that our method is quite good also in practice. 

\section{Summary}\label{sec:summ}
We have constructed a new  algorithm  for  finding a
critical  point 
of broad classes of cost functions  %defined 
on the orthogonal groups. This method is second-order-convergent  
since it  is in essence the Newton method.
The method here constructed  is an extension (or a restriction) of
the multiplicative updating method 
 developed in our 
previous work\cite{akuzawa8}. The constraint for $\Delta$ from the nature
of the orthogonal groups  makes the
problem a little complicated. We have, however, obtained a rigorous and 
explicit updating rule. 
We have also constructed 
 a Levenberg-Marquardt-type variation, which is  suitable for
 practical purpose.  
The global instability inherent in the Newton method is remedied in
this version. 
%  the Kullback-Leibler information, the kurtosis, {\it etc.},
%  suitable for
%the
%purpose. 
Since our discussion does not depend on the
detail of the cost function, 
this method is applicable to many concrete problems.
The relatively  mild assumption (\ref{eq:a1}) on the form of the  cost
function, however,  implies that 
 our algorithm is especially
suitable for 
 the ICA. 
%we can choose arbitrary functions for
%$\{f_i\}$. 
% readily our method 
%by
%prewhitening data. 
%The potential of our method  
Its practical utility for the ICA
 have been  illustrated here  by a numerical simulation.

%Let us conclude the a
To summarize, 
our algorithm  has  numerous theoretical virtues such as 
the  rigorous second order convergence, the explicit and strict formulation,
 and so on. 
%Moreover  
 It provides, 
 also in practice, 
  fast and powerful tools for the
 ICA and many other problems.

%Since it does not require prewhitening, 

\section*{Acknowledgments}
The author would like to thank Noboru Murata and Shun-ichi Amari for 
valuable
discussions and comments. 
%\bibliography{mybib}

\end{document}